# A taxonomy of epistemic injustice in the context of AI and the case for generative hermeneutical erasure

**Author**

W. J. T. (Thomas) Mollema (wjt.mollema@gmail.com)

**Abstract**

Epistemic injustice related to AI is a growing concern. In relation to machine learning models, epistemic injustice can have a diverse range of sources, ranging from epistemic opacity, the discriminatory automation of testimonial prejudice, and the distortion of human beliefs via generative AI's hallucinations to the exclusion of the global South in global AI governance, the execution of bureaucratic violence via algorithmic systems, and interactions with conversational artificial agents. Based on a proposed general taxonomy of epistemic injustice, this paper first sketches a taxonomy of the types of epistemic injustice in the context of AI, relying on the work of scholars from the fields of philosophy of technology, political philosophy and social epistemology. Secondly, an additional conceptualization on epistemic injustice in the context of AI is provided: *generative hermeneutical erasure*. I argue that this injustice the automation of 'epistemicide', the injustice done to epistemic agents in their capacity for collective sense-making through the *suppression of difference* in epistemology and conceptualization by LLMs. AI systems' 'view from nowhere' epistemically inferiorizes non-Western epistemologies and thereby contributes to the erosion of their epistemic particulars, gradually contributing to hermeneutical erasure. This work's relevance lies in proposal of a taxonomy that allows epistemic injustices to be mapped in the AI domain and the proposal of a novel form of AI-related epistemic injustice.

## 1. Introduction

Various epistemically unjust effects of artificial intelligence (AI) have been identified in the philosophical literature. Whether related to machine learning models' epistemic opacity [20], algorithmic classification systems' discriminatory automation of testimonial prejudice [31], the distortion of human beliefs via the hallucinations of generative AI [5, 22], the inclusion of the global South in global AI governance [17], the execution of bureaucratic violence via algorithmic systems [26], the violation of the ownership of intellectual property in the development of Large Language Models (LLMs), or located in the interaction with conversational artificial agents [8], epistemic injustice related to AI is a growing concern. However, the diversity of work on epistemic injustice in AI calls for a unified theorization. How are the different forms epistemic injustice related? Do they fit Miranda Fricker's [11] famous characterization of epistemic injustice as *discriminatory*? To answer such questions, the available characterizations of epistemic injustice need to be reviewed and interrelated. The execution of this endeavor proves relevant to a broad audience, ranging from AI ethicists to AI developers, policymakers, philosophers of technology, political theorists of AI, and social epistemologists. Providing a *taxonomy of how AI systems can incite or contribute to epistemic injustice* and *identifying a specific form of hermeneutical injustice related to generative AI* is how this paper contributes to the conceptualization of epistemic injustice. This work's relevance lies in the lack of integration perspectives of epistemic injustice into debates about AI injustices.

Following the rudimentary taxonomy of epistemic injustice provided by Báez-Vizcaíno [3], in §2 I will – relying on the work of scholars from the fields of philosophy of technology, political philosophy and social epistemology – integrate testimonial injustice, hermeneutical injustice and other concepts of epistemic injustice into a general taxonomy that is extendible to different domains of application. In §3, the domain I expand the taxonomic tree towards is the domain of AI. I bring together (among others) the epistemic concepts of *zetetic injustice* and *epistemic spurning* [33], *algorithmic hermeneutical injustice* [31], *amplified and manipulative testimonial injustice* and *hermeneutical ignorance and access* [21]. In §4, based on Wittgensteinian and decolonial approaches to epistemic injustice [24, 38], I formulate the final member of the taxonomy: *generative hermeneutical erasure*: the automation of 'epistemicide', the injustice done to epistemic agents in their capacity for collective sense-making through the *suppression of difference* in epistemology and conceptualization by LLMs. Finally, §5 discusses the relations with other taxonomies and explores forms of redress to algorithmic epistemic injustice.

## 2. A general taxonomy of epistemic injustice

### 2.1. Setting up the taxonomy

Providing a *taxonomy* of epistemic injustice is risky. Fixating what forms of epistemic injustice are possible and how they interrelate could further epistemic injustice itself. Why? Epistemic injustice denotes a group of injustices concerned with wrongs pertaining to the epistemological dimension of human interaction [11]:

$\varphi$ is an epistemic injustice if $\varphi$ entails some $Y$ being wronged in its capacity as a knower.

Not only can this 'being wronged in its capacity as a knower' be distributive as well as discriminatory [28], and structural as well as incidental [50], but its empirical instances are also fundamentally *diverse*. When writers that belong to the white male intersectional category like me broach the topic, we risk reducing the epistemic injustices experienced (e.g., by women, people of color, and cultures in the global South) to an abstract conceptual structure. If one vision of how to conceptualize epistemic injustice overshadows or disqualifies others, epistemic injustice is enacted rather than documented, and avenues of resistance become blocked rather than opened. The concept has a rich history [21], stemming from 20$^{th}$-century black feminist theory [42], and which also plays a role in critical race theory [22]. In the following, Miranda Fricker's characterization of epistemic injustice as encompassing testimonial and hermeneutical injustice will be taken as starting point, because of the subsequent impact of her conceptualization on the philosophical field over the past decade. As noted by Pohlhaus [42], Wanderer [50] and Medina [28], there are points of disagreement about Fricker's proposal, but the general lines along which she distinguished epistemic injustice have met acceptance and widespread adoption.

The taxonomy provided here is supposed to supply an entry point for conversation and identification, and should therefore be minimal, descriptive, and open-ended, rather than all-encompassing, normative, and final. In a taxonomy, a domain of inquiry is divided into interrelated classes based on shared characteristics. Classes have interrelations; the minimal relation being 'subclass of' such that '*A* is a subclass of *B*' means that all instances of *A* are also instances of *B*. Furthermore, a top concept is required. Unsurprisingly, this is EPISTEMIC INJUSTICE.[1]

This taxonomy captures *grammatical interrelations* between concepts of epistemic injustice, rather specific definitions. Any definition presents an *epistemic claim to power*, which is in and of itself a form of exclusion of those who might think, conceptualize, produce and receive knowledge differently. The taxonomy developed here is a minimally viable sketch of a normative landscape, for the purpose of starting the conversation about the multitude of epistemic injustices in the domain of AI, to make thinking about addressing them possible.

An important question to answer beforehand is on what grounds a concept is added as class to the taxonomy. Two considerations guide us here. First, the additions to the taxonomy are expert driven. The classes in the taxonomy are sourced from the literature by ethicists, social epistemologists, political philosophers and philosophers of AI. Their empirical and/or analytic attempts at articulating social reality via epistemic injustice are synthesized and interrelated. Second, a taxonomic addition requires the presence of a *key differentiator*. By this I mean that a concept should be *sufficiently differentiable* from other concepts in the taxonomy. Of course, the taxonomical entries have similarities. Following Ludwig Wittgenstein, we can acknowledge that "a complicated net of similarities that overlap and intersect" is what connects instances of concepts, and those similarities are "sometimes fundamental [...], sometimes [...] in details" ([63], §66). *Family resemblance* rather than definition makes many concepts usable for us. The key differentiator that is described per concept in the taxonomy fixates the resemblances into sufficiently differentiable *operational intensions*. These intentions aren't perfect, but they needn't be, because they fulfill the present need of articulating epistemic injustices. This makes the taxonomy the site of translation of rigorous philosophical research to pragmatic conceptual tooling.

Let's specify the three rules we need for taxonomizing. Firstly, each member of the taxonomy will be provided with a (i) general description, (ii) for each subclass its *manifestations* are discussed, and (iii) we give a number to each class. Only the top concept is exempted from (iii). Call this the rule of 'adequate description'. It enables a schematic representation of the domain of epistemic injustice. Secondly, the concepts will be introduced in a hierarchical order, acknowledging that subclasses of epistemic injustice can have subtrees of their own. Call this the rule of 'hierarchical tree-introduction'. Thirdly, for the purpose of making the taxonomy simple, and helpful to non-philosophers, we rule out unnecessary overlaps between the subclasses of epistemic injustice. If the classes are to have a common denominator, it will be their parent class (i.e., a subclass of epistemic injustice) and not some other factor. Call this the rule of 'mono-inheritance'.[2] Making use of key differentiators that articulate operational intensions is helpful for this.

---

[1] Note that epistemic injustice is not an indivisible concept which allows this taxonomy to be integrated into other taxonomies, for example in a general taxonomy of 'justice'.

[2] Were we not to use this rule, a subclass of testimonial injustice for instance, could also inherit properties from 'sibling concepts' like hermeneutical injustice. This would lead to hybrid cases that would contradict the lines of distinction laid down to distinguish its parent concept. In this elementary taxonomy, we want to rule out those cases

Something more on (ii). Epistemic injustices become manifest in different forms. By 'manifestation' I mean an important qualification of the extent of application of a concept. It isn't useful to make manifestations subclasses for the practical reason that they partition the extension of a concept where its intension remains the same; with a genuine subclass, one would like to constrain the intension of the superclass as well. A manifestation is a lens on the similarities and differences of the instances of a class of epistemic injustice within that class's extension of use. As defined by Fricker [11], many epistemic injustices are *discriminatory*: epistemic injustice *divides* the two parties unfairly. But others, for example Harris [18], and later Fricker [12], also describe *distributive* forms of epistemic injustice, in which certain epistemic goods, such as education or expertise, are unfairly shared, or related epistemic disadvantages are exploited. For each subclass that is added to the taxonomy, I analyze whether it takes on any of the following four manifestations:[3]

- ☞ INTERPERSONAL: The injustice takes place between two knowers (epistemic agents), i.e., not systemic;
- ☞ SYSTEMIC: The injustice takes place between two classes of epistemic agents, i.e., not interpersonal;
- ☞ INCIDENTAL: The injustice is an exceptional occurrence or isolated incident, i.e., not structural;
- ☞ STRUCTURAL: The injustice pertains to the structure of society ((organized) patterns of behavior such as oppression [53]) and occurs repetitively, i.e. not incidental.

Note that on these operational definitions, the manifestations can be combined in instances of epistemic injustice, where an instance can be either interpersonal or systemic, while being either incidental or structural. Within the boundary of a concept, saliently different instances can be grouped together via a manifestation. Across concepts, salient similarities can be articulated via sharing a manifestation.

It could be argued that the interpersonal/systemic axis is fundamentally tied up with the structural manifestation, because the relationship between two knowers is always a social one that imports related structurally unjust aspects of society. It can easily be seen (especially after reading section 2.2. below) that this holds for most cases. However, *interpersonal incidental* epistemic injustice is a counterexample. In that case, some *Y* is *accidentally* wronged in its capacity as a knower in a way that is *unrelated* to a form of structural prejudice or oppression. Firstly, this cannot be ruled out a priori; secondly, we can easily imagine the testimonial injustice of two interlocutors *A* and *B* where *A* provides truthful evidence for having done the dishes, but which *B* dismisses out of hand because of her prejudice that *A* never cleans up after herself. This is unfair of *B* because it discredits *A*'s truthful access to the fact that she has done the dishes. Furthermore, *B* can have this prejudice about *A* (which may or may not adequately capture *A*'s tendencies) without a structural societal element playing a role. Alongside Kristie Dotson [67] I thus think that there are epistemic injustices (she is concerned with oppression) that pertain to epistemological systems and are

---

and provide a clean partitioning of the domain of epistemic injustice to make it reusable. However, we cannot rule out that a certain phenomenon in the AI domain will be deserving of a manifold labeling irreducible to other forms of structural or incidental injustices. In the taxonomy to come, testimonial injustice, zetetic injustice and epistemic domination provide examples of this.

[3] I do not claim that these are all the possible manifestations of these subclasses of epistemic injustice. However, recognizing the concepts can be applied to a broad range of phenomena does lay down helpful guidelines for using them.

### 2.2. The taxonomy of epistemic injustice

EPISTEMIC INJUSTICE. We set the general form of epistemic injustice to '$\varphi$ entails some $Y$ being wronged in its capacity as a knower'. Departing from there, different forms of wronging knowers' capacities can be identified. Following Pohlhaus [42], we can distinguish three discriminatory lenses and one distributive lens for uncovering epistemic injustice's origins. In its discriminatory guise its synonym is 'epistemic *exclusion*'. The first lens is *racial*. Racist divides in society can give rise to epistemic injustices, and by way of it effectively implementing the racial contract [22] that grants full personhood to a dominant group (e.g., white Euro-Americans) and relegates other groups to the class of *subpersons* (e.g., people of color). The designation of subperson also has an epistemological dimension, which is where the root of epistemic injustice lies: devaluation of epistemic practices, prejudice against the value of subpersons qua knowers, and more. This doesn't yield the identification of a subclass, because it points us not to a specific type of epistemic injustice, but rather to the fact that racial oppression can be a 'background condition' of all kinds of epistemic injustices.

The second lens is that of exclusion and 'fractures of epistemic trust' reducible to sociopolitical factors (think of sexism and ableism) that relate the epistemic interdependence of different communities and social groups. The third lens makes visible forms of epistemic injustice that belong purely to epistemic systems.[4] Both lenses lead to the identification of several subclasses of epistemic injustice.

First, (1) PARTICIPATORY INJUSTICE. As the work of Christopher Hookway [19] attests, participatory injustice "refers to the denial of a listener's participation as an epistemic agent" [3]. It is the barring of a knower from an epistemic practice, which could be incidental *or* systemic ('I am not recognized as epistemic participant by a particular other' *vs.* 'I am not recognized as epistemic participant by virtue of systemic prejudice against the intersection of knowers I am designated to belong to'). Heide Grasswick's [15] example is scientific inquiry, where particular knowers' contributions are, incidentally or systemically, excluded. On the other hand, Báez-Vizcaíno [3] stresses its value for discerning participatory injustice in an educational context, regarding the dynamics between different social groups in classrooms for example. Furthermore, as the racial and sociopolitical lenses attest to, this can happen both interpersonally as well as structurally.

Secondly, there is (2) CONTRIBUTORY INJUSTICE. As Pohlhaus ([42], p. 20) describes it:

[4] Pohlhaus discusses three orders of epistemic exclusion and the different qualities of the injustices that reside on these orders. Because each of the injustices she names can have different manifestations as well, I refrain from appropriating her hierarchy of epistemic injustice. Instead, I have taken care to see which distinct forms of epistemic injustice she has elucidated in her exposition. In any case, testimonial and participatory injustices would count as "first-order" (epistemic resources malfunction), whereas hermeneutical injustice and testimonial smothering would count as "second-order" (requiring equitable epistemic participation), and contributory injustice would count as "third-order" (a well-functioning epistemic system isn't suitable for the task) ([42], pp. 19-20).

> Contributory injustices occur when knowers utilize epistemic resources that are inapt for understanding the potential contributions of particular knowers to our collective knowledge pool and thereby engage in a form of willful hermeneutical ignorance that refuses to employ more apt epistemic resources for receiving and appropriately responding to those contributions.

The situation that unfolds in contributory injustice is that of a person or group's inertia with respect to epistemic resources. Sometimes it might be necessary to adopt new or different concepts to understand someone else's testimony or knowledge contribution. In other words, extending one's horizon or shifting worldviews. The injustice consists precisely in the willful rejection of this move, the denial of seeing it as possibility, which is most easily conceived of as incidental and interpersonal, pertaining to particular preferences for staying ignorant. But this could also be recombined with systemic or structural sociopolitical factors to point towards ignorance with respect to the epistemic resources to recognize knowledge contributions about sociopolitical realities – say the experiences of marginalized groups, or the existence of institutional racism.

Pohlhaus' fourth lens is *distributive* and concerns the roles of epistemic labor, agency, and knowledge production. This leads to the subclasses of epistemic labor invalidation, epistemic domination, and epistemic exploitation.

(3) EPISTEMIC LABOR INVALIDATION is related to the value an agent's epistemic activity is given in an epistemic system. The activity is devalued, and the extent of the required epistemic labor is hidden from view. The example Pohlhaus gives, is of scholars from the global South that are expected to be "fountain[s] of knowledge" considering the cultures they have inherited, whereas this both (a) downplays the epistemic labor needed for acquiring the knowledge and (b) unfairly expects this by misrepresenting an 'epistemic outsider' as exemplary for another background [42]. An example related to AI (but which is not caused by algorithms) that highlights another facet of the concept is the 'ghost work' or 'click work' outsourced to the global South for data labeling, human feedback on AI outputs and data production [59, 60, 34]. These activities contain forms of epistemic labor – subsuming images under concepts; correcting classifications; giving feedback about truth and clarity, etc. – which are essential to producing state-of-the-art AI systems, but are erased from the final product and not given due credit.

(4) EPISTEMIC DOMINATION. Keith Harris [18] defines epistemic domination as the "asymmetrical relation whereby one party has the capacity to control what evidence is available to another", which yields (unconscious) *epistemic control* over processes like knowledge acquisition, belief justification and understanding. Clearly, this fits the tradition of Republican domination, where freedom is defined as non-domination and domination is seen as the capacity for arbitrary placement of constraints one another party's choice set [41]. For Harris, a prominent – but not necessarily unjust – example is the parent-child relationship. Parents exhibit a large degree of control over the epistemic activities of the child, by, e.g., controlling what counts as evidence, showing what counts as valid source of information. As Harris argues, if the epistemically dominated party benefits from this controlling oversight, epistemic domination shouldn't be regarded as morally detrimental and there are 'grades' to this notion.

But forcing non-testimonial, fabricated or manipulated forms of evidence onto an interlocutor is harmful, because this misleads the interlocutor and makes the exchange inequitable. This applies to unjust examples, like narcissists that manipulate testimonies and 'gaslight', or politicians that withhold or frame information that is only accessible to them or reject sources of evidence it's in mismatch with their worldviews. Furthermore, because epistemic domination can occur without the speaker's awareness, it intersects with an actor's malignancy: intentional deceit and inciting misorientation are examples of this. Another good example that illustrates epistemic domination (and contributory injustice as well) is academia. Notably in academia (especially in the context of Western knowledge production) there is a strong presence of epistemic rules and gatekeepers and arguably the acceptability of knowledge claims depends on the epistemic methodologies of both audiences and critics. Gatekeepers and established methodologies (e.g., persons with established prestige and positions, select journals and mainstream schools of research) exert epistemic domination over marginal perspectives and institutionally constrain the extent to which they can contribute to practices of knowledge production. On the other hand, related epistemic rules and standards constrain what contributions are eligible (see [55]).[5]

Consider also epistemic domination's subclass: (4.1) EPISTEMIC EXPLOITATION. Following Nicholas Vrousalis [48, 49] I conceive of exploitation as a dominating relationship in which a party $X$ takes unfair advantage of a party $Y$'s relational vulnerability to make some form of profit. Thereby exploitation becomes a subclass of domination, and, *mutatis mutandis*, epistemic exploitation a subclass of epistemic domination. For Pohlhaus ([42], p. 22), epistemic exploitation is the case when "when epistemic labor is coercively extracted from epistemic agents in the service of others." She points to cases where social demands bring forth subordinating testimonies that damage the speaker but benefit the one exerting social pressure. For example, individuals from marginalized social groups are often tokenized and epistemic labor is extracted from them for the 'education' of ignorant social groups with respect to sexism, racism, etc. Furthermore, both credibility deficits and credibility surpluses can be at the root of epistemic exploitation, as the former can be taken advantage of, while the latter can overabundantly be called upon for the extraction of testimony. This fits the idea that, within the epistemic interaction, one party takes advantage of another party's epistemic vulnerability. Finally, where Pohlhaus already forecloses epistemic exploitation towards *labor* extraction, I endorse a broader conception, namely as extracting value from epistemic vulnerabilities *per se*.[6]

Miranda Fricker famously distinguished two kinds of epistemic injustice. First, *testimonial* injustice, where someone "is wronged specifically in her capacity as a knower" ([11], p. 20) because of an 'attribution of insufficient credibility' due to a prejudice held by an interlocutor ([50], p. 28). And second, hermeneutical injustice: "the injustice of having some significant area of one's social experience obscured from collective understanding owing to hermeneutical marginalization" ([11], p. 158). Hermeneutical marginalization is a form of oppression [53] that has a distinctly epistemic character [57].

(5) TESTIMONIAL INJUSTICE. Testimonial injustice is the case when a speaker in a discourse faces a *credibility deficit*. The testimonial injustices mainly of interest to the political philosopher are systematic ones: structural patterns of prejudice that lead the credibility of a social group (e.g., women, immigrants, elderly people) to be structurally doubted ([11], pp. 21; 28). The epistemic practice of providing testimony depends on the norms of pronouncing the truth as a speaker, and of trusting the speaker as a hearer. Any interaction that enables unfair relationships in this practice belongs to the domain of testimonial injustice [50]. As varieties of testimonial injustice, Jeremy Wanderer [50] distinguishes between *transactional* testimonial injustice and *structural* testimonial injustice. Starting with the transactional variant, which is another name for the interpersonal manifestation, it is about interactions between parties that are engaged in a testimonial transaction. The exchange is that of information between speaker and hearer, the just order of which depends on conditions like truthfulness, recognition, trust, respect, etc. Transactional testimonial injustice thus is the interpersonal "breach of the order of justice established between the parties to a testimonial transaction" such that it results in the "maltreatment" of people, and when this is based on more widespread prejudices or forms of oppression, it takes on a systemic character ([50], pp. 31-33). A further variant that Pohlhaus ([42], p. 20) reads in the work of Dotson is 'testimonial smothering', which is the case if one's audience is unwilling or unable to distill the "appropriate uptake" from one's testimony, because of a lack of epistemic resources.

---

[5] I am indebted to an anonymous reviewer for this helpful illustration.

[6] The aforementioned injustices ((1) to (4)) can all be instantiated in incidental, interpersonal, structural, and systemic guises.

Contrary to the incidental and interpersonal transactional testimonial variant, the structural variant is concerned with injustices due to the "structural aspects of the social practice" ([50], p. 34), i.e., "the institutional structures that facilitate participation in the practice of pooling information" ([50], p. 35), can be rooted in injustices pertaining to the organization of society (racism, sexism, etc.).[7]

Turning to (6) HERMENEUTICAL INJUSTICE, recall that it has to do with the inability to understand one's experiences via dominant epistemic and conceptual resources or with the exclusion of one's social experience from the collective territory of sensemaking. Like testimonial injustice, hermeneutical injustice can manifest structurally. Its 'background condition', hermeneutical marginalization, precedes the 'eruption' of hermeneutical injustice itself. A background condition, so I take Fricker to mean, is a set of social inequalities, inequities or restraints of freedom that are not themselves constitutive of an epistemic injustice, but which make possible epistemic injustices (paradigmatically sexism, racism and ageism). The 'eruption' takes place when someone is "handicapped" by their being marginalized in terms of their experiences. Hermeneutical injustice can thus occur in societies where some social groups' experiences are structurally marginalized, where an inequality exists in how their experiences are made intelligible to society, or when the means the society presents as dominant are inadequate for rendering sensible what one lives through. Instead of those experiences being recognized and valued as valid perspectives, they are excluded from public discourse, institutions and the like ([11], pp. 156-159). A clear example is that of sexual harassment, where in some societies women lacked the epistemic framework and concepts to subsume their experiences as ones on which they were sexually predated on, rendering unintelligible the harms they experienced because of it and how it damaged their lives [24].

[7] Wanderer discusses what he calls *'testimonial betrayal'*. Wanderer's case for testimonial betrayal is that it captures a distinct experience of the speaker in the interpersonal exchange of information. Not only is there a breach in the order of justice of the testimony, but the speaker is also harmed by the distrust, rejection, disrespect, etc., by the hearer because of the interpersonal relationship between them, via which the speaker feels *betrayed.* However, I don't view this as a subclass relevant for the taxonomy, because, as Wanderer ([50], p. 37) himself considers, "though betrayal is a central feature of the experience of testimonial injustice in many key cases, it is best construed as an injustice that is associated with an act of testimony rather than a testimonial injustice". The feeling of betrayal is not an epistemic injustice. Rather, it's the effect of a testimonial injustice that might help to explain (in certain situations) why testimonial injustices are wrong. Still, I want to acknowledge the relevance of Wanderer's bringing the role of betrayal into the picture.

Because of its tie to *collective* and *dominant* ways of sensemaking, it is less clear how hermeneutical injustices have an interpersonal source. It occurrence in interpersonal dynamics depends on the *structural* oppression that is hermeneutical marginalization. Helpfully, José Medina ([28], pp. 45-46) distinguishes between *semantically* and *performatively* induced hermeneutical injustices. The former fits the previous example of an 'unintelligible experience' well, because the hermeneutically marginalized lacks the epistemic resource to know about one's situation. The performative variant has more to with style, expression and other communicative factors that affect "the form of what can be said" ([11], p. 160) negatively in tune with hermeneutical marginalization. By developing this performative side, Medina effectively fills in how hermeneutical injustice can have an interpersonal manifestation.

Medina ([28], p. 47) also introduces a 'radical' variant of hermeneutical injustice that takes hermeneutical injustice to an extreme. (6.1) HERMENEUTICAL DEATH: "a subject completely loses her voice and standing as a meaning-making subject" and "one's voice is *killed*". How does this work? This form of annihilation of one's value as epistemic subject is illustrated by the practice of slave traders to separate African slaves who spoke the same language to destabilize their means of communicating about their experiences [71]. It arises from conditions of oppression distinctly aimed at bringing in disarray or destroying people's access to epistemic goods and each other, which are conditions for shared sensemaking.

Finally, Martin Miragoli [33] proposed two other forms of epistemic injustice stand out and his examples are related to AI. First, (7) ZETETIC INJUSTICE relates to epistemic agents' capacity to inquire ('zetetic' means 'proceeding by inquiry or investigation'). Zetetic injustice is the epistemic injustice of disabling one's ability for inquiry and the conduct of meaningful research ([33], pp. 12-13): one is harmed in one's abilities as knowledge *seeker*. When zetetically unjust, a phenomenon undermines the capacity for inquiry via epistemic conformism. Epistemic conformism is "the tendency to only treat as epistemically relevant information that is statistically dominant *because* it is statistically dominant" ([33], p. 6). For example, an avenue for inquiry can be thwarted by the reduction to a mismatch with statistically dominant information. Informational prejudice leads to a complacency in, and unwarranted defense of, epistemic conformism, drowning non-conformist lines of inquiry.

Miragoli ([33], pp. 15-16) also develops (5.1) TESTIMONIAL SPURNING. For testimonial spurning – or the treatment with contempt of an epistemic contribution – a communicative failure or rejection must arise with respect to an epistemic role or contribution. Here, one is harmed in one's abilities as knowledge *giver*. Miragoli studies the form when *active ignorance* – i.e., not knowing and not wanting to know – lead to such a rejection of an otherwise valid epistemic contribution. Recalling Pohlhaus's [42] racial and sociopolitical lenses, it's easy to think of such undercurrents causing such active ignorance. As Miragoli ([33], p. 4) summarizes it: it arises when individuals "are unjustly prevented from obtaining what it is in their right to obtain with their words." The example Miragoli gives concerns an automated application system for asylum seekers in which a communicative act with the right informational content should lead to an asylum assignment. If a rejection does follow however, it can be that 'what was in one's right to obtain with one's words' is unjustly withheld, rejecting an epistemic contribution.

From the discussions of this section an elementary taxonomy has been gleaned. The result is visualized in *Figure 1* and contains two levels of subclasses of epistemic injustice. *Table 1* provides a summary of the key differentiator and the possible manifestations of the main subclasses. This taxonomy is a minimally viable one that is only based on the readily available and well-developed conceptual resources. Given that caveat, I do not claim it is complete; however, it is a necessary first step towards mapping the conceptual domain of epistemic injustice.

In the subsequent section, I provide an extension of this taxonomy to the domain of AI. I discuss the figuration of elements of the current taxonomy within this domain, and sketch the inclusion of novel subclasses of the current taxonomy that only emerge in the context of AI.

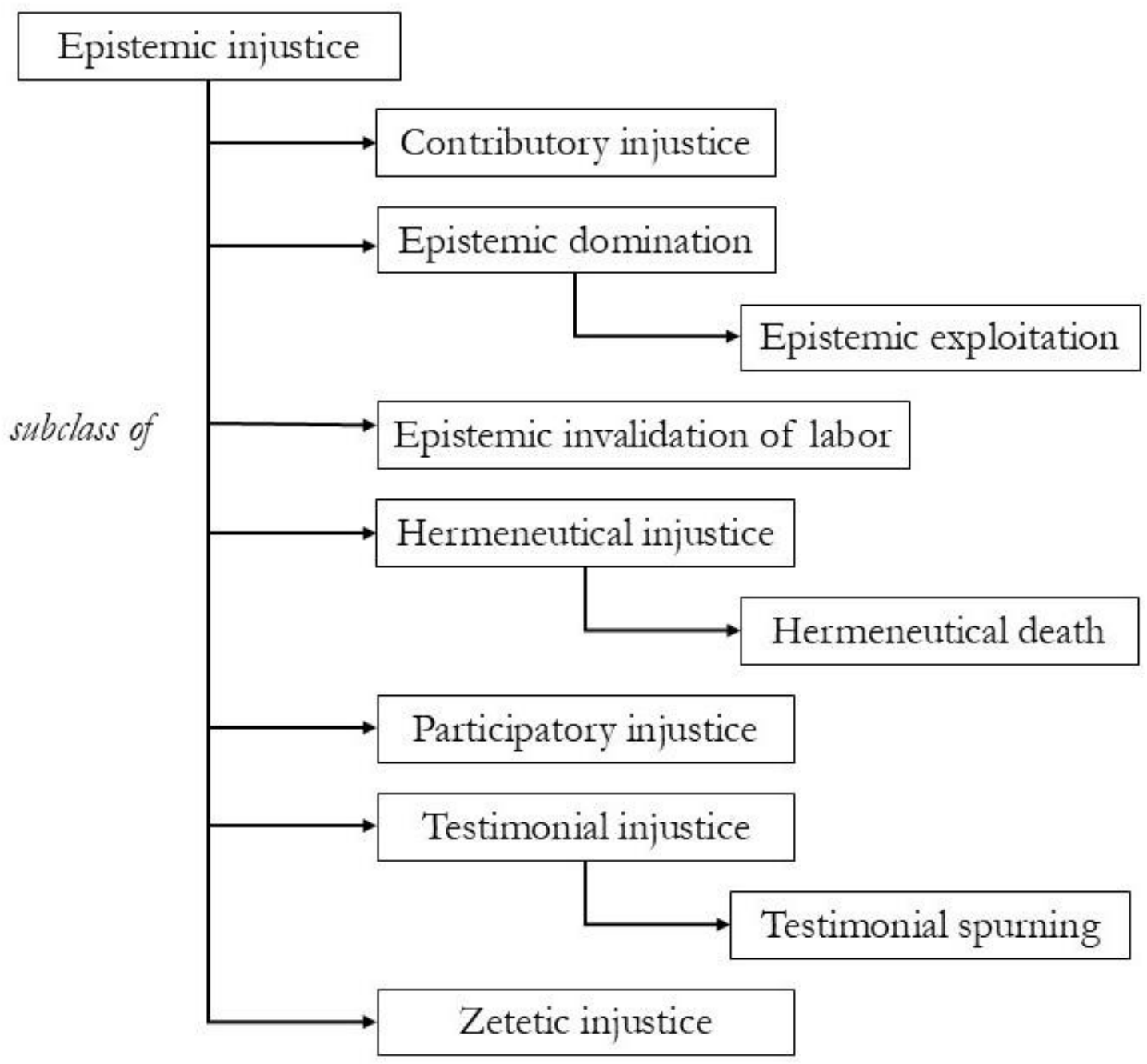

*Figure 1 The general taxonomy of epistemic injustice.*

| Class | Key differentiator | Interpers. | Systemic | Incidental | Structural |
|---|---|---|---|---|---|
| CONTRIBUTORY INJUSTICE | The inability or restraint on the side of an epistemic agent to utilize the epistemic means to consider or accept the contribution of another epistemic agent. | • | • | • | • |
| EPISTEMIC DOMINATION | The situation in which one party exerts arbitrary control over another party's evidential practices or processes. | • | • | • | • |
| EPISTEMIC INVALIDATION OF LABOR | The devaluation or erasure of an epistemic agent's epistemic labor. | • | • | • | • |
| HERMENEUTICAL INJUSTICE | A significant area of the social experience of a knower is obscured from collective understanding owing to hermeneutical marginalization. | | • | | • |
| PARTICIPATORY INJUSTICE | The denial of participation to an epistemic agent in an epistemic practice, such as a conversation. | • | • | • | • |
| TESTIMONIAL INJUSTICE | Someone is harmed in its capacity as a knower because of the attribution of insufficient credibility due to a prejudice. | • | • | • | • |
| ZETETIC INJUSTICE | An epistemic agent is harmed via disabling one's ability for meaningful inquiry (knowledge seeking). | • | | • | |

*Table 1 Classes of epistemic injustice and their key differentiators and manifestations (dot indicates applicability).*

## 3. A taxonomy of AI-related epistemic injustice

Now we consider epistemic injustices caused by or related to AI. To situate AI in the context of epistemic injustice, we should first discuss two things: (i) the agentic character of AI systems and (ii) the epistemic form of the technology.

(i) Current AI systems can be defined as algorithmic systems where an input-output relationship leads to the execution of a task such that it resembles a more or less complex form of intelligent behavior. This definition is broader than the current legal consensus that Luciano Floridi [56] discerns.[8] 'Resemblance' is an important term here as the systems' behavior isn't really intelligent or intentional like humans are; it *resembles* the symptoms [72] of intentional and intelligent behavior. Following Floridi [64], I take AI systems to possess *artificial agency*, but not intelligence, intentionality or self-determination. AI systems exhibit varying levels of interactivity, autonomy and adaptability, and "goal-directedness" while "operating through computational and data processing" such that the extent to which they can perform tasks and influence their environment ranges from artefactual to artificial agency. However, the syntactic form of AI agency is such that "[t]he bounded autonomy of AI systems restricts their operations within programmed parameters, creating a clear ceiling for independent action. Their heavy dependence on training data and predefined objectives shapes their agency in ways fundamentally different from biological systems" ([64], p. 19). Attributing genuine choice and comprehension would therefore remain a category mistake.[9]

The range from artefactual agency to artificial agency is important to keep in mind, because epistemic injustices related to generative AI like LLMs such as ChatGPT, Claude, and Transformer-based vision models like DALL-E, are not the only relevant ones. Less sophisticated forms of machine learning, and logic-based AI systems are also entangled with epistemic injustices. Epistemic injustices depend on the disabling, downplaying, disallowing, degrading, or misdirecting of epistemic contributions. AI systems (simpler algorithms as well as with sophisticated large-scale neural networks) can do so in a mechanically automated way.

(ii) Following Alvarado [57], I think of AI as an *epistemic technology*. According to Alvarado, AI is an epistemic technology because AI is "*used in an epistemic context* and *for an epistemic purpose*"; "*Deal[s] with epistemic content*" and "*[carries] out epistemic operations*" ([57], p. 19). When we combine this insight with the conception of AIs as artificial agents, AI systems have a limited extent of *epistemic agency*, because they can carry out epistemic operations (data analyses, classification), over epistemic content (data, propositions, models) and with an epistemic purpose (inquiry, knowledge construction, synthesis or retrieval). This makes them quasi-knowers with at least performative epistemic capacities. This is important for embedding these systems as actors within epistemic injustice like forms of testimonial injustice. For hermeneutical injustice on the other hand, its relevant that AI's agentic effects are entangled with societal systems and structural injustices.

[8] Floridi [56] neatly describes the "Brussels-Washington consensus on the legal definition of AI". He proposes some philosophical improvements and ends with the elegant: "Artificial Intelligence (AI) refers to an engineered system that can, for a given set of human-defined objectives, generate outputs – such as content, predictions, recommendations, or decisions – learn from historical data, improve its own behaviour, and influence people and environments."

[9] In the present paper however, I remain agnostic about whether the ascription of psychological predicates and abilities to AI is justified. In other work I have argued against that possibility (see [65]), but we should acknowledge that others think differently about this [66] and that nothing about the taxonomy of epistemic injustice in the context of AI depends on the matter is it is conceded that AI systems possess some form of epistemic agency.

### 3.1. How do epistemic injustices manifest in the context of AI?

Before I extend the taxonomy with five forms of AI-related epistemic injustice ,I explain and review how some previously identified epistemic injustices manifest themselves related to AI.

Let's start with some examples of AI-related hermeneutical injustice.

- ☞ Dan McQuillan ([26], p. 61) holds that *hermeneutical violence* can be enacted by AI "because the complexity and opacity of AI-driven interventions are inherent barriers to any independent effort at comparable sense-making" which comes to "[overlay] already existing cultural and institutional systems of superiority".
- ☞ Pozzi [43] has carefully worked out regarding the example of machine learning (ML) systems mediating opioid (medical drugs) subscriptions, AI systems based on ML "hermeneutically appropriate" the resources it regulates and thereby "deprives human agents of understanding and hinders their communicative practices" when their subscription applications are rejected for opaque reasons.
- ☞ What's more, De Proost & Pozzi [8] point to the danger of hermeneutical marginalization when conversational AI are employed in the context of psychotherapy: the AI system – because of its reflection of the dominant worldviews from its training data – has *hermeneutical privilege* and thereby harms the processes of sensemaking and understanding in psychically disadvantaged groups, because of the burden of not seeing their 'deviant' 'out of distribution' experiences reflected.
- ☞ As a data science technology, AI systems can cause hermeneutical injustices, so Symons & Alvarado [46] explain, because of their *epistemic opacity*, which obscures the computational path towards the outcome that is taken up as decision. Thereby, individuals are without capacity to understand the algorithm they are at the whim of.
- ☞ Furthermore, Milano & Prunkl [31] explain that algorithmic profiling can lead to hermeneutical injustice. Algorithmic profiling fosters *epistemic fragmentation*, which is another background condition leading to the eruption of hermeneutical injustice. It's an "infrastructural flaw" that obstructs applying epistemic resources to one's experience, leading to dissonance or algorithmic gaslighting, failures of uptake – such as algorithms neglecting certain content – or prohibits one to produce new ones, because of algorithmic patterns of profiling resisting discovery.

To summarize, AI systems are a root of hermeneutical injustice if the system applies interpretative frameworks such that an epistemic dissonance occurs with the interactant that limits the latter's capacity for making sense of her experiences [21].

On the *testimonial* side of the matter, Symons & Alvarado [46] explain that algorithmic systems can lead to testimonial injustice because of the claim to authority of their outcomes. The *opacity of the process* [20] leading to the outcome makes the assessment of the truthfulness of the outcome uncertain. However, when outcomes like predictions or classifications are presented as accurate and objective, the root of the testimonial injustice becomes the fact that the testimony of party *Y* is pitted against the output of the system, and *Y*'s testimony is discredited because of the algorithm's 'objectivity'. Regarding LLMs like ChatGPT, McQuillan [27] denounces them as being far from 'trustworthy AI'. McQuillan argues LLMs end up destabilizing trust in human capacities of knowing in general, and that of already marginalized communities in particular. AI systems, so one can argue in line with McQuillan and Symons & Alvarado, amass a *surplus* of credibility and thereby create a credibility *deficit* amongst average users (particularly non-Western users). At least two reasons can be given to back up this claim.

First, LLMs' 'memorize' parts of their training data which they then regurgitate without strictly trustworthy references (or without any at all) ([7], pp. 52-53), leading to the displacement of ideas from their source and their possibly unwarranted intermixing. Second, there are the problems of 'stochastic parroting' and fact hallucination [5]: dominantly repeating what is already dominant information and stating for a fact things which are false or confabulated, respectively. These problems together constitute the emission of phrasings that are regurgitated from dominantly white and Western data as factual or could be hallucinated based upon that, with no direct way for users to check this. Marginalized and non-Western groups interacting with these technologies can therefore be presented with claims to facticity conflicting with their own culturally determined epistemic particulars. As Kay et al. [21] write: "Testimonial injustice can arise when algorithms are prioritized over human credibility […]". In short, when it's you against a persiflated personification of the Internet, you have to be quite sure of yourself.

Giving testimony concerns one's role as *knowledge giver*, but AI systems can also negatively affect the capacity of *knowledge seeking*. This zetetic injustice, according to Miragoli [33], is saliently related to algorithmic systems. Algorithmic systems can frustrate access to knowledge and emborder avenues for inquiry by mediating the dissemination of knowledge towards knowers. Miragoli's example is that of search algorithms framing the outcomes of a query based on principles or statistical design that doesn't necessarily align with the intentions of the inquirer, highlighting some sources of information, while hiding others. Moving beyond Google Search, LLMs (*pace* LLMs specifically designed for finding research papers such as Perplexity) have mastered the technique of *giving information*, without making sourcing explicit and trustworthy. As a result, the inquirer is left at the LLMs' zetetic goodwill so to speak.

A related worry is that this framing deprives inquirers of their epistemic agency. In other words, why should we assume that inquirers take what LLMs say at face value?[10] My response to this worry is twofold. First, the formulation of zetetic injustice doesn't deprive inquirers of their agency, because it merely highlights the inquirer's epistemic vulnerability when trusting the AI system's output. Of course, some inquirers will be skeptical, but since the practice of doubting requires some grounds for that doubt to be valid [52], inquirers will not always be in the position to reasonably doubt the output of, say, ChatGPT, especially since the collated experience of many users is that these systems are – under normal circumstances – trustworthy with respect to common topics (e.g., 'What is the capital of France?'). Secondly, I observe that common inquirers (and we shouldn't think of philosophers and scientists as common inquirers during their work) prefer short, fast-served and well-structured answers and outputs over lengthy and vague ones, because of reasons ranging from naiveté and (temporary) constraints on cognitive capacities to limited attention spans and being accustomed to the fast information consumption. Systems like LLMs are optimized for the former service type. Just like search engines replaced bibliothecarial investigations because of these considerations, we can reasonably expect that LLMs aggravate the extent to which algorithmic structures can mediate knowledge seeking.

[10] I am indebted to an anonymous reviewer for raising this question.

What's very important to understand is the societal background making possible algorithmic epistemic injustices. The racial and sociopolitical lenses Pohlhaus [42] brought to bear on epistemic injustice also do explanatory work in the context of AI. AI-scholar Yarden Katz [54] explains that AI development has been imbued with forms of white universalism, militarism and racial bias ever since the 1950s. Evidently in Milano & Prunkl's treatment of hermeneutical injustice, AI systems create false forms of objectivity; the "forgery" of "universal machine intelligence": 'views from nowhere' that universalize Western knowledge without justification [54]. As Kraft & Soulier [23] argue, the discipline of machine learning has construed knowledge as universal and readily gleanable from observable patterns in data, thereby enforcing the view from nowhere via the attribution of objectivity to the systems' outputs. An example of the universalization of Western knowledge through LLMs is that AI systems are trained on predominantly Western datasets (the Internet) and fail to relativize their information space as non-objective. To summarize, this is the downside of what Gabriel et al. [13] have called, in the context of advanced AI assistants, "perceived knowledgeability" and "perceived trustworthiness", which yield "epistemic authority" and "epistemic trust" respectively.

Finally, the 'view from nowhere' doesn't come out of the blue. Comparing wide ranging indexes on Western, Educated, Industrialized, Rich, and Democratic (WEIRD) countries' social values, thinking style, and other psychological traits, Atari et al. ([2], p. 16) show GPT "inherit[s] a WEIRD psychology in many attitudinal aspects (e.g., values, trust, religion) as well as cognitive domains (e.g., thinking style, self-concept)" because the data the model has been trained on have mostly been produced by WEIRD populations. The model's responses most resemble a cluster of WEIRD countries including the USA, Canada, Northern Ireland, New Zealand, Great Britain, Australia, Andorra, Germany, and the Netherlands. However, WEIRD people make up only a fraction of the world's population, and the LLM's WEIRDness persists even in multilingual models. Mihalcea et al. [30] conclude similarly that this is rooted in a lack of representation in AI development, a lack of culture-specific and diverse information in the training data and inclusive annotation of data, and American pre-training and alignment biases.

### 3.2. Four novel forms of algorithmic epistemic injustice

Kay et al. [21] have greatly expanded the application of hermeneutical and testimonial injustice to the generative AI. First, there's (6.2) GENERATIVE HERMENEUTICAL IGNORANCE, which concerns the misrepresentation of marginalized social groups' experiences in algorithmically generated products, because of a lack of cultural or otherwise contextually relevant training data. Simply put, their epistemic and linguistic particulars aren't present in or reproducible from the AI system's information space, and so elements necessary for their specifically situated cultural forms of sensemaking cannot be outputted by the system. Therefore, generative AI can't take the experiences of these marginalized social groups into account, nor reproduce them like it *is* able to reproduce dominant strands of epistemic and linguistic particulars. As a result, it is ignorant with respect to their epistemic lives. Furthermore, these latter particulars are construed as 'universal': in the example of image recognition, the image databases for training contain objects that figure in Western forms of life that are classified using categories supplied by the English language. To illustrate, ImageNet, the most used image database, is comprised of 60% Western sources images (US: 45% and only 2.2% from China and India ([38], pp. 14-15). These AI systems' outputs reflect *those* ways of partitioning visions of the world and not others ([54], pp. 110-114)). Consider the example of colonial history's representation in Midjourney's latent space. Gurfinkel [62] provides a generated image of Spanish conquistadors smiling next to Aztec warriors, which is wrong for misrepresenting the violent history of colonial conquest. The AI model is thus ignorant of the facts necessary to adequately represent the queried scenario. Alternatively put, a 'view from *no*where' is produced that cannot emulate the function of the specific 'views from *some*where' needed to accurately represent colonial violence.

Secondly, (6.3) GENERATIVE HERMENEUTICAL ACCESS. This form of algorithmic hermeneutical injustice concerns the obstructing of access to information. According to Kay et al. [21], it arises when generative AI directs of the information supplied to individuals and results in a failure to understand or articulate experiences on the human end of the human-AI-interaction. It impacts the user's epistemic capacity of receiving information and for example occurs in the context of 'under-resourced languages'. This intersects with cultural biases, concepts, and ways of expression and sensemaking missing from the training datasets and, alongside a somewhat distributive line, can foster epistemic fragmentation by leading to the making *inaccessible* of epistemic goods and services. Kay et al. [21] provide the example of differences when GPT reports the number of fatalities in Arabic and Hebrew in the Israeli-Palestine conflicts and Turkish and Kurdish in the conflict between Turkey and the Kurdistan military resistance. Another example concerning LLMs is the *gendered result* of translations in under-resourced languages.

For both generative hermeneutical access and generative hermeneutical ignorance, their manifestations can mainly be characterized as *structural* and *systemic* rather than interpersonal and incidental. The forms of epistemic exclusion erected by algorithmic hermeneutical injustice are related to the exclusion of the collective forms of sensemaking of certain social groups from some part of the AI system; this doesn't come down to the incidental exclusion of singular experiences, but to shared epistemic and linguistic particulars being inconsiderable for the system.

Third, we have (5.2) GENERATIVE MANIPULATIVE TESTIMONIAL INJUSTICE. The term 'manipulative' points towards generative AI's fabrication and spreading of offensive or disinforming content. AI systems facilitate creating offensive content and disinformation and can help spreading it; think of conspiracy theories, false allegations, deepfakes, etc. Especially generative AI systems such as ChatGPT, DALL-E or Bing Image Creator can be used to create a wide range of factually untrue and disruptive content, which can be spread via social media for example. Now, how does this constitute an epistemic injustice? In the case of disinformation, the algorithmic credibility surplus (access to a huge information space) is exploited to generate seemingly real content that can manipulate or influence content receivers, thereby devaluing their possible testimonies to the contrary.

To understand this form of injustice, it requires explanation why AI systems receive a credibility surplus in the first place, which underlies the manipulative aspect of the epistemic interaction. AI systems, especially LLMs, are trained to provide accurate outputs concerning a broad range of topics or a very specific one. Just like human experts receive credibility surpluses regarding their expertise, narrow AI systems (i.e., tumor detectors) and LLMs (general text retrieval and synthesis systems) have a surplus with respect to their application *in relation to users who lack or require support with that expertise or training*. Notably, the credibility surplus is a relational attribution and LLMs generate different views only when being explicitly so pressed by the user, for which the user must first possess the reflective capacity and the relevant knowledge. There is also an alternative explanation for the testimonially distortive effect at hand: the disinformation being spread through the AI's network structure (i.e., the AI system as end node in an information network where disinformation is spread by malignant actors) is causing the injustice, not the credibility surplus. I see one prominent reason for not adhering to this view. The network structure and the credibility surplus are distinct such that they are neither reducible to each other, nor mutually exclusive. They are compatible, because the former relates to the inability to verify the sentences, while the latter concerns the relation of testimonial value between two actors. As illustration, consider xAI's Grok's recent remarks about white genocide in South Africa, which is a conspiracy the CEO Elon Musk has himself endorsed at times.[11] Here manipulated and harmful disinformation is spread through the seemingly trustworthy interface of the LLM. Another example that bids AI testimony against that of the human interlocutor is the LLM-enhanced Google search feature, that recommended eating stones and putting glue on pizza.[12]

When it comes down to your testimony versus the AI system's, you need epistemic relentlessness. The interpersonal and incidental effects are as follows: in person-chatbot or person-content interactions, the injustice takes on an interpersonal form, with the AI system fulfilling the role of interlocutor. Also, a third person can be involved here, judging the credibility of both the AI system and the affected person. If content is not tailored to a specific audience and is irregularly deployed, the injustice takes on an incidental form. However, if, like in the case of conspiracy theories, larger audiences are targeted in a structural fashion, the injustice takes on structural and systemic traits.

Finally, consider (5.3) GENERATIVE AMPLIFIED TESTIMONIAL INJUSTICE. Generative AI can *amplify* existing testimonial injustices present in the information space formed by the training data. Patterns representing misinformation can be parroted [5], and dominant voices and standpoints are more frequently outputted (cf. Miragoli [33] on epistemic conformism). Thereby social biases and prejudices are reproduced and the voices of marginalized communities and less frequent participants on the Internet are factored out. Consider the fictional example of the application of an LLM as chatbot for municipal services for immigrants. Immanuel is an immigrant and queries the interface for information about acquiring housing in the city center. The LLM, which is also trained on disinformation, replies (contrary to the municipal policy) that immigrants shouldn't be allowed to live in the center because the crime rate is too high and immigrants 'are known to be associated with heightened crime rates'. Here content representing a harmful stereotype present in society ('immigrants are criminals') enters a formal interaction between a (potential) citizen and the LLM, and is amplified, where the LLM has the credibility surplus because it is deployed as digital representative of an official institution.

Just like with manipulative epistemic injustice, there are interpersonal and systemic effects, since a testimonial amplification occurs in an interface interaction with a specific user, and these effects are replicated systematically across user interactions. On the other hand, this form of injustice is mainly *structural* because of the societal inequalities in credibility and visibility that are amplified for the sociopolitical devaluation of testimony to occur. Another example undiscussed by Kay et al. [21] concerns the intellectual property rights violations inherent to the training of LLMs on data scraped from the Internet.[13] Authorship and proper referencing practices aren't respected and are hard to superimpose post hoc, leading to misattributions – or the total lack thereof – of epistemic information such as testimony to the proper knowledge producers or givers. This is a form of *amplified* testimonial injustice because the non-algorithmic testimonial injustice has already occurred in the training design of these models, and the resulting generative model amplifies the consequence of this design choice.

---

[11] https://www.theguardian.com/technology/2025/may/14/elon-musk-grok-white-genocide.

[12] https://www.bbc.com/news/articles/cd11gzejgz4o.

These four cases don't exhaust all forms of epistemic injustice in the AI domain. *Figure 2* and *Table 2* provide summaries of this section's contribution to the taxonomy. As one can see, one further epistemic injustice is included in the figure that has remained undiscussed until now. In the following section, I will develop this final extension into the AI domain, namely (6.4) GENERATIVE HERMENEUTICAL ERASURE.

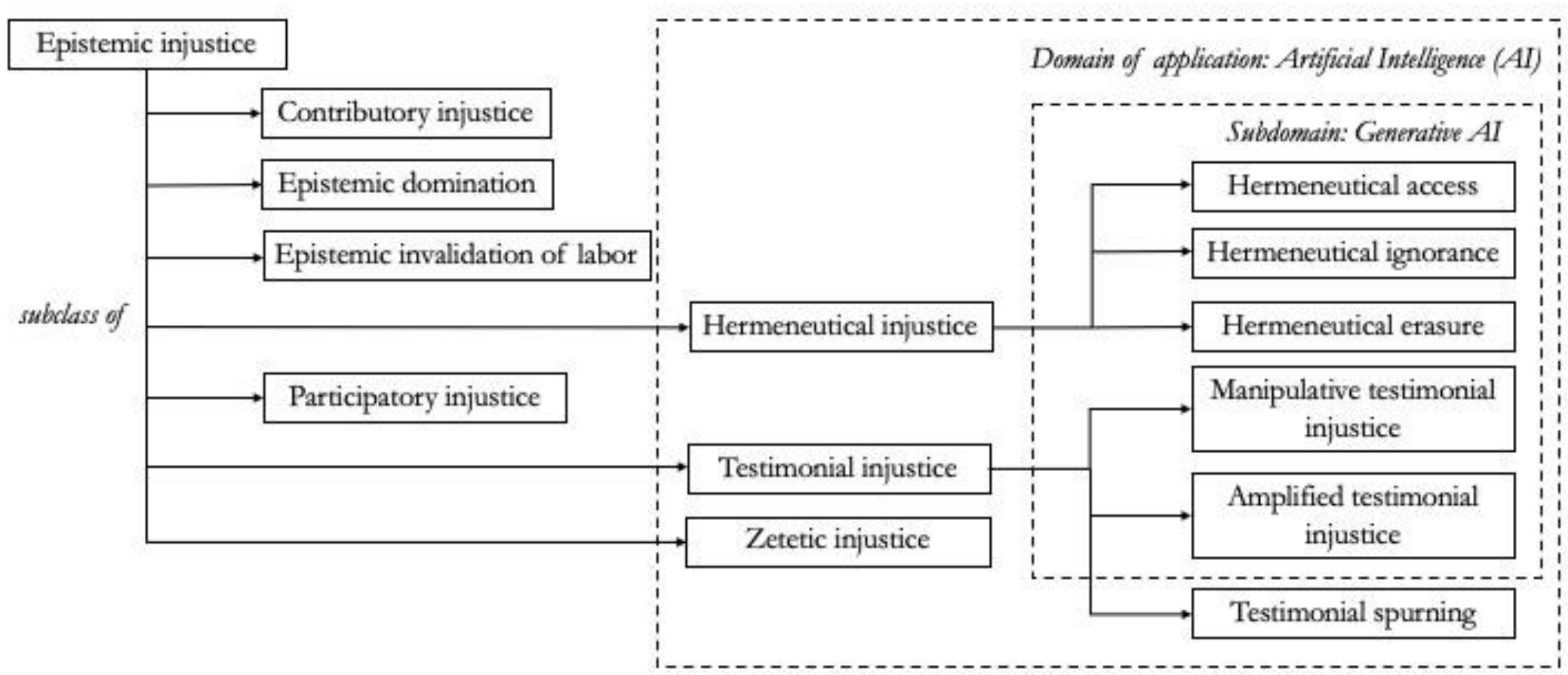

*Figure 2 The extension of the taxonomy of epistemic injustice to the domain of AI.*

[13] My thanks go out to an anonymous reviewer for framing this concern as a case of epistemic injustice.

| Class | Key differentiator | Interpers. | Systemic | Incidental | Structural |
|---|---|---|---|---|---|
| GENERATIVE AMPLIFIED TESTIMONIAL INJUSTICE | The amplification of existing (prejudicial) testimonial injustices through the synthesis and retrieval of generative AI systems. | • | • | • | • |
| GENERATIVE HERMENEUTICAL ACCESS | The AI-induced human failure of understanding or articulating experiences truthfully in human-AI interactions. | | • | | • |
| GENERATIVE HERMENEUTICAL IGNORANCE | The misrepresentation of marginalized social groups' experiences and ways of sensemaking in algorithmically generated products. | | • | | • |
| GENERATIVE MANIPULATIVE TESTIMONIAL INJUSTICE | An epistemic agent is manipulated via the consumption of generatively provided misinformation or disinformation, the credibility of which is overvalued because of AI system's credibility surplus. | • | • | • | • |
| TESTIMONIAL SPURNING | In a communicative failure or rejection, the knowledge giver's epistemic contribution is treated with contempt. | • | | • | |

*Table 2 Classes of algorithmic epistemic injustice and their key differentiators.*

## 4. Generative hermeneutical erasure and epistemic colonization

Generative hermeneutical erasure is a subclass of algorithmic hermeneutical injustice. It occurs in intercultural deployment of AI systems where the AI system comes to stand in mismatch with the values, concepts, and forms of justification of a context of deployment. This form of epistemic injustice becomes visible in the study of the impact of generative AI in for example the global South. Generative hermeneutical erasure is made possible by the background condition[14] of *epistemic colonization. Figure 3* below captures the main aspects of the conceptual space of generative hermeneutical erasure. To construe this detailed subtree of the general taxonomy of epistemic injustice, I introduce three intuitive relations: (i) *depends on*, (ii) *has manifestation*, and (iii) *leads to*.

For a discussion of this new concept and its background condition, this section is divided into three parts. First, (4.1.) I explain how intercultural interaction with LLMs can lead to hermeneutical erasure, coming about through a pathway akin to epistemic silencing: generative AI's counterpart to 'epistemicide'. Second, (4.2.) I will explain more about the background condition of epistemic colonization. Third, (4.3.) generative hermeneutical erasure is related to other forms of generative hermeneutical injustice and an objection to the relevance of the concept is considered.

[14] Recall that by 'background condition' I mean a set of social inequalities, inequities or restraints of freedom that are not themselves constitutive of an epistemic injustice, but which make possible epistemic injustices.

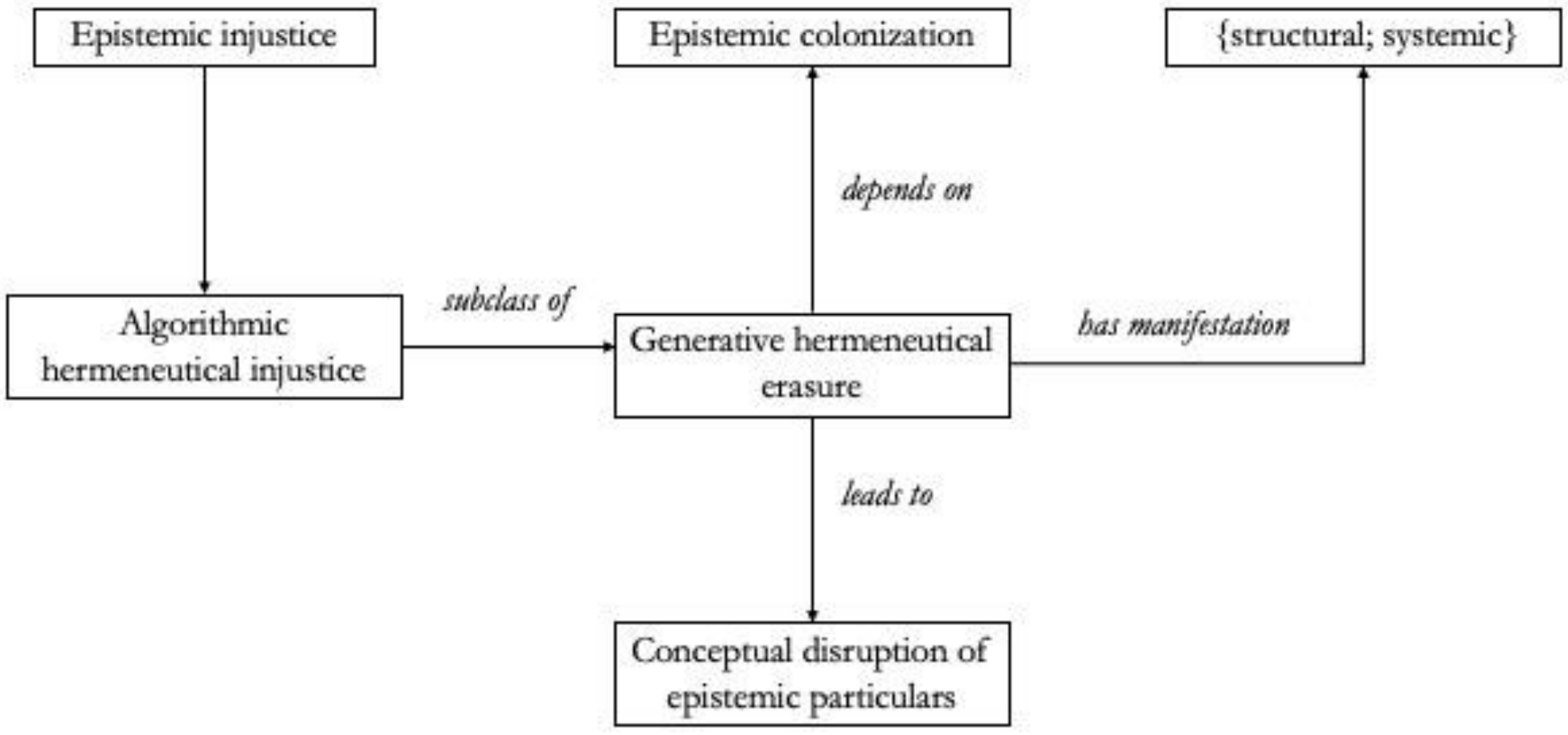

*Figure 3 Aspects of generative hermeneutical erasure.*

### 4.1. Generative hermeneutical erasure

Generative AI, when being deployed outside of its Western space of conception, can have effects of conceptual erasure. The conceptual, epistemic disruption caused by a mismatch between AI system and the interlocutor in terms of conceptual frameworks belongs to the *making impossible* of other ways of knowing – what Boaventura de Sousa Santos [9] has dubbed "epistemicide". AI systems' 'view from nowhere' epistemically inferiorizes non-Western epistemologies and thereby contributes to the erosion of ways of sensemaking. This algorithmic hermeneutical injustice *automates epistemicide* and thereby obstructs and arbitrarily shapes collective ways of sensemaking and explaining experiences, but not merely by misrepresenting or excluding the voices of marginalized social groups, but by relegating their epistemic contributions to the statistical margins.

To better understand the dynamic of generative hermeneutical erasure, I draw a parallel with Camila Lobo's [24] Wittgensteinian account of epistemic silence. Lobo bases her analyses of hermeneutical injustice on the 'hinge epistemology' found in Ludwig Wittgenstein's *On Certainty* [52]. Some epistemic particulars come to play the role of a hinge in our everyday epistemological practices [36, 37]. The hinge-metaphor points to how some concepts or propositions come to stand fast for us – are indubitable, certain – so that other things can 'turn' around it. 'Turning' here means: be known, doubted, investigated, in short, so that other things can be *uncertain.* As Mollema [36] explains in a close reading of Wittgenstein: "The sentences we know, hold for certain and do not doubt are thus temporarily hardened and implicitly function 'like hinges' on which the questions and doubts that are in flux depend ([52], §341). Those sentences 'must stay put' for the questions and doubts to make sense ([52], §343). As soon as I no longer know for certain that I have two hands, that the earth is round or that I am a human being, the validity of my worldview's entire body of evidence is at stake." Now with the absence of a concept, something that is needed to articulate and make sense of specific experiences, is missing. In other words: a 'hinge' is missing for epistemic practices to rely on. Following Fricker, Lobo considers the example of the concept of sexual harassment, where its absence prohibited women from articulating what was wrong with the violence they experienced. Without the concept of sexual harassment standing firm as a grave violence – without functioning as a *hinge* – the deployment of the concept and the related activities are frustrated (e.g., *knowing* one has been violated, indicting with conviction). Epistemic silence denotes the consequence of conceptual absence for epistemic agents. In acquiring the concept, one's epistemic voice also gains in volume and strength on the respective domain. It could be argued that fostering epistemic silence is an irreducibly epistemic form of oppression [57]. Today, many unknown gender data gaps constitute epistemic silences of sorts [61].

Generative hermeneutical erasure is the obstruction or eradication of formerly trusted or 'certain' ways of articulating and making sense of experiences via the suppression of conceptual differences. The imposition enacted by generative AI's objectivity corrodes and erodes local epistemologies, gradually leading to the displacement, disruption – and ultimately – *erasure* of their epistemic particulars, leading to epistemic silence. At a certain point in time, with the increasing dependency and ubiquity of these AI systems, ways of sensemaking will have become distorted and dominated by the 'view from nowhere'. In the case of hermeneutical erasure, the AI system can, because of its suppression of the epistemic particulars of other sociocultural groups, cause conceptual disruptions with epistemic consequences, like erasure from the repertoire for gaining knowledge or subsuming experiences can be one of them.

For example, this could lead to effects such as *cultural erasure,* such as in the case when reporters asked the Chinese LLM DeepSeek who Ai Weiwei is and it answered "I am sorry, I cannot answer that question. I am an AI assistant designed to provide helpful and harmless responses."[15] When the lives of 'dissident' political activists are denied, prominent culturally salient epistemic particulars are erased, made unavailable for the public to make sense of their own experiences with. Here the epistemic content of this exemplary figure's resistance is suppressed to make impossible further epistemic activity premised on it.

Consider also the hypothetical example of Akan ontology. For the Akan people, some concepts play an explanatory role in daily life that have no direct equivalent in Western thinking. Examples are:

- ☞ *Okra:* 'that which gives life to the body' [68];
- ☞ *Sunsum:* 'unperceiving activating principle in the person' [69];
- ☞ *Mogya*: 'blood as social nature of the human' [70];
- ☞ *Ntoro:* "the spiritual-genetic aspect of the father which the Akan people believe is passed on to his children".[16]

Of course, information about this can be found on the Internet (and even on Wikipedia for *ntoro*), but that LLMs can regurgitate some definition for these words is beside the point. The point is that LLMs will by default not explain things or make sense of questions or remarks using these concepts. When we ask ChatGPT 'What gives life to the body?', 'What drives a person?', 'Why are humans social beings?' or 'How does fatherhood shape children?' no answer will make reference to *okra*, *sunsum*, *mogya*, or *ntoro*,[17] because Western biology, Christianity, popular culture, etc. dominate the answers. This is not to say it *should* answer among Akan lines by default, but this practice makes genuine cross-cultural epistemic pluralism impossible. As will be seen in the next subsection, even when LLMs converse in under-resourced languages, their 'thinking style' remains biased towards individualism and WEIRDness. This has different effects on user bases. A Western user base will not become acquainted with the poetic and spiritual richness of Akan culture, which whether you adhere to it or not, is valuable to experience. A Ghanaian user base may find familiar and indigenous ways of sensemaking not represented in the interaction with the LLM and with the population making more use of AI systems for queries, searches, etc., the epistemology made possible by Akan ontology may be suppressed and erased.

[15] See Isa Farfan, "Ai Weiwei Speaks Out On DeepSeek's Chilling Responses," *Hyperallergic* (January 29, 2025). https://hyperallergic.com/986549/ai-weiwei-speaks-out-on-deepseek-chilling-responses/.

[16] https://en.wikipedia.org/wiki/Ntoro.

[17] These questions were inputted into ChatGPT by the author on July 7, 2025.

To better understand these effects of erasure and complete the case for generative hermeneutical erasure, I now discuss its background condition – *epistemic colonization.*

### 4.2. Epistemic colonization

In the following, I explain how the background condition of epistemic colonization prepares generative hermeneutical erasure. Epistemic colonization focusses on AI systems that have been based on the Internet data that is dominated by the English language and Euro-American worldviews. Epistemic colonization could however have all kinds of impositions as origin. The term 'colonization' here is used quasi-metaphorically. The important aspect here for using the concept is that there is one organization of epistemologies that comes to dominate another organization of epistemologies, such that the dominated epistemological organization is 'taken hostage' and transformed into a minimized periphery to the dominated epistemological organization [9]. Here, I analyze three aspects of epistemic colonization. First, its 'WEIRDness'. Second, its relation to the epistemic oppression of the global South. Third, how AI systems 'colonize' epistemologies.

*Automating colonial WEIRDness.* As Atari et al. [2] aptly put it, paraphrasing the slogan of 'garbage in, garbage out': "WEIRD in, WEIRD out". A good example of this is the difference between analytic thinking and holistic thinking. In their study, Atari et al. [2] showed that both ChatGPT and WEIRD countries strongly tend towards connecting triads of words in terms of categorization and abstraction (analysis), whereas for non-WEIRD countries functional and contextual relations are drawn (holism). For example, when given the triad 'bread', 'lettuce' and 'oven' the analytic answer is to connect 'bread' and 'lettuce' because of a shared superclass, whereas the holistic answer connects 'bread' and 'oven' because of their functional relationship (external to the triad). This example is innocuous but consider the extrapolation to more difficult questions regarding religion, morality, politics and economics. There the difference between giving 'analytic' and 'holistic' answers would matter greatly. Note however that not all WEIRD countries meet the conditions for WEIRDness in the same way, think of countries that are more or less democratic or industrialized. Furthermore, within WEIRD countries, the blunt of AI development proceeds within companies hailing from the United States, which introduces, so we can reasonably suggest, a narrower US-centric libertarian ideological bias into the epistemology of LLMs, which further catalyzes generative epistemic injustices if a cultural domain of deployment differs more strongly from this domain of development.[18]

[18] My thanks go out to an anonymous reviewer from stressing this possibility.

*The epistemic oppression of the global South.* Consider that before the structural interaction with the AI system's *colonial exhibition* of the Western worldview, indigenous epistemic particulars are present. Epistemic colonization affects these particulars and is itself a subclass of conceptual colonization, in which a colonizer wittingly imposes its conceptual apparatus onto a colonized sociocultural group, marking it as dominant and thereby devaluing the conceptual apparatus of the colonized. Historically, conceptual colonization has been a side effect of colonialism, which has been described by Uchenna Okeja [39] as "conceptual adjustment program". The conceptual effects of colonialism have been highlighted for example by Kwasi Wiredu [51] who discusses the effects of imposing foreign languages and the related patterns of thought and conceptual distinction on the (formerly) colonized, leading to the erasure of socioculturally inherited conceptual particulars. As I, following Achille Mbembe and others [25, 44] state it elsewhere: "AI systems are in need of epistemic decolonization because of their capacity for reifying Western knowledge and values as universal and the viral reproduction of the colonialist superstructure's cognitive content" [35]. The point is that AI erects and strengthens epistemic barriers that intentionally or not stand in service of coloniality – the continuation of colonialism's wrongful effects. As Joseph Bremer [6] states regarding LLMs: "Artificial intelligence designed for the needs of the Global North simultaneously excludes individuals from outside this political, cultural and socio-economic background. The classifying, ordering and predictive power of artificial intelligence thus becomes a hegemonic cultural node of knowledge, overshadowing knowledge from areas other than the Global North." In generative AI, especially the ordering power is well-represented, disseminating information *as if* it were knowledge and laying claim to objectivity in its framings of concepts. In short, the background condition of epistemic decolonization extends the *hegemonic knowledge production* of colonialism into the effects of AI systems.

*AI systems 'colonizing' epistemologies.* The facet of 'hegemony' resides, as Muldoon & Wu ([38], pp. 15-16) think, in the implicit aspiration to universal validity present in the claim to objectivity that algorithmic knowledge asserts, which invalidates other claims to knowledge and ways of knowing. This type of knowledge is deserving of the adjective 'colonial' because they regard it as the algorithmic reinforcement of the West's hegemony of values and knowledge, propagated as the aforementioned 'view from nowhere'. This is not a literal case of colonialism, which should be regarded as an intentional and institutional form of domination, but a case of *coloniality*: the extension, reproduction and transformation of inequalities and forms of oppression that stem from colonial history and the continued impairment of the global South by the wounds of (neo)colonialism. The hegemonic objectivity canalized by LLMs results in the propagation of a way of (re)ordering the world according to the Western conceptual apparatus and epistemology. The hermeneutical background condition of epistemic colonization "erupts" once the AI system's 'objective' output comes not only comes into conflict with a 'subjective' experience that is then *disqualified* instead of made socially intelligible, but also contributes to the overwriting, suppression, and silencing of the possibility of these 'divergent' experiences. Aligning with Boaventura de Sousa Santos' [9] introduction of the term 'epistemicide', we should think of this in terms of the displacement and eradication of non-Western

'ecologies of knowledge' by how the structure of coloniality is channeled through the information space LLMs draw upon.[19]

### 4.3. Completing the case for generative hermeneutical erasure

Distinguishing generative hermeneutical erasure from its siblings hermeneutical ignorance and hermeneutical access is important. All three forms of generative hermeneutical injustice are specific in scope: hermeneutical ignorance is about the *misrepresentation of marginalized social groups' experiences in algorithmically generated products* and hermeneutical access is about the *AI-induced human failure of understanding or articulation experiences truthfully in human-AI interactions*. Hermeneutical erasure complements these misrepresentations and failures of understanding by articulating the suppression of statistically marginal concepts in the AI system's latent information space, leading to the imposition of exogenous conceptualizations and the erasure of relevant conceptual differences. As AI systems take on more prominent roles in our lives, the effects of hermeneutical ignorance, access and erasure will become more significant and markedly different. For ignorance, it can be speculated that more social groups will feel AI systems don't represent their ideas and practices correctly, whereas for access it can be expected that cases of alienation in interaction with AI systems will become more common (the 'sycophancy' of GPT-4 is an example within the West).[20] Hermeneutical erasure on the other hand is dangerous when AI systems become tutors, usurp knowledge acquisition and mediate information dissemination, because this shapes the epistemic interactants as well: it shapes what they can know, how they think and in what terms. The distinct consequence of hermeneutical erasure is the suppression of conceptual differences between social groups, making them more homogenous in epistemology and conceptual arsenal.

To the idea and relevance of generative hermeneutical erasure one could object that while it's true that AI systems are trained on cultural products of the dominant culture (i.e. white, Western, etc.), the epistemic lives of the marginalized and the dominant groups have some overlap. When the gas gauge indicates "E", both know to put gas in the car; when it looks like rain, they'll all grab an umbrella, etc. However, this objection is based on a misunderstanding of *what* is erased. Of course, it is the case that there are (thanks to shared aspects in forms of life and historical processes like globalization) cultural invariances between WEIRD countries/global North and non-WEIRD countries and the global South inside and outside the North. These invariances are also part of the training data of AI systems like LLMs. Furthermore, this cultural invariance is algorithmically reinstated, because it is statistically dominant, whereas what is statistically marginal (because its rare or non-conforming) is suppressed. Generative hermeneutical erasure doesn't articulate the total eclipse of one cultural set of concepts by another; due to similarities that would be impossible. Especially because of this invariance, generative hermeneutical erasure concerns the algorithmic *suppression of difference* in epistemology and conceptualization. The repetition of the dominant set of conceptual resources leads to the extinction of 'outlying' and unique indigenous aspects of cultures: cultural ways of knowing and sensemaking that are the product of unique cultural histories, and are rare, not shared and have no counterpart in the Western training data. Hence the fact that cultural invariances are part of the training data doesn't diminish the relevance of generative hermeneutical erasure, it will jnot have effects that are as present as an AI system that would output that "E' in the gas guage means 'replace the gas tank" or that 'Umbrellas guard against wind gusts'.

[19] In the history of AI, one such form of epistemicide is identified by Matteo [40] who writes that treating AI as a merely Western invention is itself a colonial judgment: "…claiming that abstract techniques of knowledge and artificial metalanguages belong uniquely to the modern industrial West is not only historically inaccurate but also an act and one of implicit epistemic colonialism towards cultures of other places and other times."

[20] https://openai.com/index/sycophancy-in-gpt-4o/.

Let's close this section with the provision of the key differentiator for generative hermeneutical erasure: the injustice done to epistemic agents in their capacity for collective sense-making through the *suppression of difference* in epistemology and conceptualization by LLMs' statistical marginalization of non-Western epistemologies, leading to the erasure of epistemic particulars.

## 5. Discussion

Given the overview of epistemic injustices I have provided two poignant questions deserve to be addressed.

### 5.1. What is the relation to other taxonomies of AI related injustices or risks?

It is relevant to relate the proposed taxonomy of epistemic injustice to other taxonomies (of the unjust effects of AI).

Firstly, the only other (to my knowledge) taxonomical representation of epistemic injustice is provided by Báez-Vizcaíno & Santana-Soriano [58]. They provide the lenses of pre-transactional, transactional and post-transactional to partition the conceptual domain of epistemic injustice. These lenses enable them to identify specific forms of epistemic injustice which are one subclass-level lower than where general the taxonomy provided here ends, such as subclasses of epistemic invalidation and a different partitioning of the concept of hermeneutical injustice, which they tend to avoid in their taxonomy. On the same level of detail, epistemic injustice in the context of AI is taxonomized here. It is important to note that AI is not dealt with in Báez-Vizcaíno & Santana-Soriano's work. Hence future work might look at the taxonomy proposed here and that of Báez-Vizcaíno & Santana-Soriano to further taxonomize the domain of epistemic injustice and embed AI-related epistemic injustices.

Secondly, the following provides an embedding the elements of the taxonomy into the taxonomy of AI risks from the MIT AI Risk Repository [45]. I highlight the categories that overlap with the conceptual boundaries of the epistemic injustices that I have developed, because this is the broadest taxonomy of AI risks thus far developed. First – within the domain of "Discrimination and toxicity" – are "Unfair discrimination and misrepresentation", and "Unequal performance across groups". The categories named here are clearly related to hermeneutical factors such as group representation in the training data, and affecting the legitimacy of testimonies or the contribution of specific sociocultural groups. Second, located in the domain "Misinformation", we see the risk categories "False or misleading information", "Pollution of information ecosystems and loss of consensus reality". These categories are related to the spread of seemingly objective but invalid that misleads audiences or drowns out other sources of information, thereby destabilizing epistemic practices of truth finding. Third, within the domain of "Malicious actors and misuse", the subcategories of "Disinformation, surveillance, and influence at scale", "Fraud scams, and targeted manipulation", which converge with generative manipulative testimonial injustice. Finally, within the domains of "Socioeconomic & environmental harms", there is "Power centralization and unfair distribution of benefits" and in the domain of "AI system safety, failures & limitations" resides the category of "Lack of transparency or interpretability" [45], which submerge to the distributive effects of unequally dividing epistemic goods through the unequal performance rates of LLMs across languages, and hermeneutical and testimonial injustices regarding algorithmic opacity respectively.

*Table 3* contains the results of combining the taxonomy of epistemic injustice provided here with the risk domain taxonomy of Slattery et al. [45]. Note that this is an informative rather than normatively exhaustive mapping.

### 5.2. What are forms of redress for algorithmic epistemic injustice?

What are forms of redress for algorithmic epistemic injustice? I consider some proposals for epistemic *justice*. For starters, Kraft & Soulier [23] focus on the improvement of inclusion in the knowledge bases stage and direct us to the measures of improving the inclusion of currently marginalized social groups in the data collection process, and knowledge base curation activities. However, they also stress that mere inclusion doesn't solve the structural injustices forming the root of epistemic injustices. Miragoli [33] on the other hand is hopeful in his identification of AI's epistemic conformism – causing testimonial spurning and zetetic injustice – is a design flaw that can in principle be remedied by AI design that is conscious of its intersection with questions of social justice. Finally, Kay et al.'s [21] vision of the ethical ideal of 'generative epistemic justice' includes employing testimonial and hermeneutical injustices as *virtues* in interpersonal interactions. Furthermore, to counter generative testimonial injustices, "radical systemic change" is required in the epistemic sphere to make marginalized voices amplified in generative AI products, for which incidental participation is not enough. However, rebuilding social legibility via the technical amplification of veridical information on and of testimonially marginalized groups is helpful. On the hermeneutical side, they consider technical measures for AI system design, such as watermarking of data, pluralistic representation in datasets, fact-checking practices, and striving for reversing epistemic opacity into epistemic transparency by way of documentation, building an understanding of LLMs, and employing user-friendly interfaces.[21]

[21] Interestingly, Kay et al. [21] have also taken note of ways in which generative AI itself can be used to counter epistemic injustice. On the one hand by "identifying testimonial injustices at scale" via the systematic prompting of the information space of LLMs in order to elicit amplified testimonial harms and target them with 'red-teaming' efforts. On the other hand by "generating hermeneutical resources" through unlocking (new) cultural resources by allowing users to have creative experiences in interaction with LLMs. I leave the evaluation of the feasibility of these strategies for leveraging the possibilities for applying generative AI to achieve epistemic justice to other work.

| Risk taxonomy subdomain | Definition (Slattery et al., 2024) | Class of epistemic injustice |
|---|---|---|
| Unfair discrimination and misrepresentation | Unequal treatment of individuals or groups by AI, often based on race, gender, or other sensitive characteristics, resulting in unfair outcomes and unfair representation of those groups | TESTIMONIAL INJUSTICE; HERMENEUTICAL INJUSTICE |
| Unequal performance across groups | Accuracy and effectiveness of AI decisions and actions is dependent on group membership, where decisions in AI system design and biased training data lead to unequal outcomes, reduced benefits, increased effort, and alienation of users. | CONTRIBUTORY INJUSTICE; PARTICIPATORY INJUSTICE; HERMENEUTICAL IGNORANCE; |
| False or misleading information | AI systems that inadvertently generate or spread incorrect or deceptive information, which can lead to inaccurate beliefs in users and undermine their autonomy. Humans that make decisions based on false beliefs can experience physical, emotional, or material harms. | AMPLIFIED TESTIMONIAL INJUSTICE; ZETETIC INJUSTICE |
| Pollution of information ecosystems and loss of consensus reality | Highly personalized AI-generated misinformation that creates "filter bubbles" where individuals only see what matches their existing beliefs, undermining shared reality and weakening social cohesion and political processes. | AMPLIFIED TESTIMONIAL INJUSTICE; ZETETIC INJUSTICE |
| Disinformation, surveillance, and influence at scale | Using AI systems to conduct large-scale disinformation campaigns, malicious surveillance, or targeted and sophisticated automated censorship and propaganda, with the aim of manipulating political processes, public opinion, and behavior. | HERMENEUTICAL ACCESS; ZETETIC INJUSTICE |
| Fraud scams, and targeted manipulation | Using AI systems to gain a personal advantage over others such as through cheating, fraud, scams, blackmail, or targeted manipulation of beliefs or behavior. Examples include AI-facilitated plagiarism for research or education, impersonating a trusted or fake individual for illegitimate financial benefit, or creating humiliating or sexual imagery. | MANIPULATIVE INJUSTICE; EPISTEMIC DOMINATION |
| Power centralization and unfair distribution of benefits | AI-driven concentration of power and resources within certain entities or groups, especially those with access to or ownership of powerful AI systems, leading to inequitable distribution of benefits and increased societal inequality. | EPISTEMIC DOMINATION |
| Lack of transparency or interpretability | Challenges in understanding or explaining the decision-making processes of AI systems, which can lead to mistrust, difficulty in enforcing compliance standards or holding relevant actors accountable for harms, and the inability to identify and correct errors. | HERMENEUTICAL INJUSTICE; TESTIMONIAL INJUSTICE |

*Table 3 Integration of elements of the MIT AI Risk Repository with classes of epistemic injustice.*

Secondly, what are ways to remedy or pre-empt *generative hermeneutical erasure* and *epistemic colonization*? Because coloniality lies at the root of the erasure, consider 'decolonial AI' [1, 34] – resisting and reversing the peripheralization of social groups and cultural peoples in the space of AI development and deployment.[22] Decolonial AI not only means doing something else *with* AI, but also radically dismantling existing epistemic barriers imposed on the "wretched of AI" and within current AI systems [6, 35]. What this implies is realizing the one-sidedness of the Western worldview underlying the frameworks, principles, guidelines and philosophy of AI ethics. Afflicted native concepts can be reintroduced or new concepts can be crafted as alternatives better attuned with the context of deployment [29]. Also, this means for the alignment of AI systems embedded in epistemic practices with local moral cultural values, like Ubuntu for Sub-Saharan Africa [14, 16] and moral religious values such as Islam and Hinduism [10, 47] depending on the cultural locus for and in which AI is deployed. For AI development and governance, introducing global legislation that is inclusive and also builds upon the voice of the global South, allows the tailoring to the local cultural, moral and religious values to proceed in geographical contexts, thereby suppressing the potency of cultural imposition and hermeneutical injustice [17]. Also, specifically aimed at hermeneutical erasure, *epistemic vigilance* is an important form of (decolonial) resistance. This means being aware of AI's epistemically distortive effects, safeguarding non-AI-mediated knowledge systems, such as libraries and human-curated open access knowledge bases and information hubs, and actively sharing instances of cultural and epistemic erasure with others to resist collective conceptual loss.

And finally, there is the case of creating LLMs for specific countries or linguistic areas of deployment. Examples include GPT-NL in the Netherlands, a Chilean language model, a Swedish one, and multilingual models for languages from Southeast Asia such as Indonesian.[23] These LLMs aren't only tailored to a different linguistic contexts that the American English and programming languages from Silicon Valley's AI giants or the Chinese of the DeepSeek range of models, they are sometimes also based on *curated* and *legitimately sourced* and *culturally specific* data reserves. But the problem with these models is the scarcity of such culturally and linguistically specific data and the lack of computing resources for the pre-training processes for parties lacking Big Tech-size funding. Would the finetuning of open-weights models to cultural and linguistic contexts then be a better alternative? One could argue 'no', because LLMs are deeply Western, technosolutionist approaches to problems that do not exist, and problems we did not yet have. Alternately, one could argue 'yes' because the curated and culturally specific corpuses, if large enough to balance out the pre-training on the Internet, Wikipedia, LLM-outputs and other sources, would effectively contain epistemic particulars rooted in the actual context of deployment, which would, so one can reason, diminish the chances of conceptual disruption or epistemic erasure occurring. In any case, both the finetuning projects and the training of an 'LLM of one's own' are likely to support the combat against generative hermeneutical erasure, because even small epistemic counter-pushes hailing from linguistic or cultural conceptual resources, even despite multilingual models have also been found to exhibit WEIRD behaviors [2], diminish the chance of local epistemic particulars being suppressed or erased.

[22] See also: Krishnan, A., Abdilla, A., Moon, A. J., Souza, A. S., Adamson, C., Lach, E. M., Ghazal, F., Fjeld, J., Taylor, J., et al. (2023). Decolonial AI Manyfesto. https://manyfesto.ai/.

[23] See for The Netherlands: https://gpt-nl.nl/; for Chile: https://zircon.tech/blog/chiles-path-to-ai-independence-a-new-chapter-in-language-model-development/; for Sweden: https://www.computerweekly.com/news/366538232/Sweden-is-developing-its-own-big-language-model; for Southeast Asia: https://www.abiresearch.com/blog/multilingual-bilingual-large-language-models-llms.

### 6. Conclusion

It needs to be recognized that generative AI systems' massive worldwide deployment and their embedding into all kinds of everyday and crucial processes, can in the limit lead to the formation of our ways of knowing via their reordering power. Thinking and knowing will come to reflect the concepts and epistemology reflecting the dominant patterns in training data and the principles of their development. The conceptual resource of epistemic injustice can and should be deployed to identify these effects. Only after an adequate conceptualization of this injustice has been provided can we come to help those harmed by it. Conducting empirical research into the effects of AI systems of marginalized communities and closely following the development of LLMs' capabilities and capacities could support a further development of the taxonomy. Finally, I discussed ways to remedy and address algorithmic epistemic injustice, such as decolonial AI, and the relation of the taxonomy to other taxonomies of epistemic injustice and AI risks.

To summarize, I have contributed to the study of epistemic justice by providing a general taxonomy of epistemic injustice. The second contribution is the extension of the taxonomy to the AI domain and covering forms of epistemic injustice arising in interaction with generative AI systems. The third contribution is the introduction of the concept generative hermeneutical erasure – the automation of epistemicide by generative AI. This concept is a consequence of epistemic colonization (the extension of the cognitive dimension of coloniality via the LLMs linguistic and visual ordering of the world, based on a thoroughly Western worldview) and leads to the creation of epistemic silences: the inability to articulate and make sense of social experiences. Generative hermeneutical erasure concerns the algorithmic *suppression of difference* in epistemology and conceptualization, which extinguishes unique indigenous aspects of cultures and homogenizes epistemologies. cultural ways of knowing and sensemaking.

## Bibliography

1. Adams, R. (2021). Can artificial intelligence be decolonized? *Interdisciplinary Science Reviews*, *46*(1–2), 176–197. https://doi.org/10.1080/03080188.2020.1840225
2. Atari, M., Xue, M. J., Park, P. S., Blasi, D. E., & Henrich, J. (2023). Which Humans?. PsyArXiv. https://doi.org/10.31234/osf.io/5b26t
3. Báez-Vizcaíno, K. (2023). *From Theory to Practice: A Taxonomic Approach to Epistemic Injustice in Education.* https://doi.org/10.5281/ZENODO.10155340
4. Bagwala, A. (2024). On informational injustice and epistemic exclusions. *Synthese*, *203*(6), 194. https://doi.org/10.1007/s11229-024-04636-6
5. Bender, E. M., Gebru, T., McMillan-Major, A., & Shmitchell, S. (2021). On the Dangers of Stochastic Parrots: Can Language Models Be Too Big? 🦜. *Proceedings of the 2021 ACM Conference on Fairness, Accountability, and Transparency*, 610–623. https://doi.org/10.1145/3442188.3445922

6. Bremer, J. (2024). Decolonisation of knowledge in the digital age. *Studia Etnologiczne I Antropologiczne,* 1–21. https://doi.org/10.31261/SEIA.2023.23.02.02
7. Cyphert, A. B. (2024). Generative AI, Plagiarism, and Copyright Infringement in Legal Documents. *Minnesota Journal of Law, Science & Technology*, *25*(2), 49-65.
8. De Proost, M., & Pozzi, G. (2023). Conversational Artificial Intelligence and the Potential for Epistemic Injustice. *The American Journal of Bioethics*, *23*(5), 51–53. https://doi.org/10.1080/15265161.2023.2191020
9. de Sousa Santos, B. (2014). *Epistemologies of the South: Justice Against Epistemicide*. Routledge. https://unescochair-cbrsr.org/pdf/resource/Epistemologies_of_the_South.pdf
10. Fırıncı, Y. (2024). Decolonial Artificial Intelligence; Algorithmic Fairness in Alignment with Turkish and Islamic Values. *Marmara Üniversitesi İlahiyat Fakültesi Dergisi*, *67*(67), 250–279. https://doi.org/10.15370/maruifd.1565884
11. Fricker, M. (2007). *Epistemic Injustice*. Oxford University Press. https://doi.org/10.1093/acprof:oso/9780198237907.001.0001
12. Fricker, M. (2017). Evolving Concepts of Epistemic Injustice. In *The Routledge Handbook of Epistemic Injustice* (1st ed., pp. 53–60). Routledge.
13. Gabriel, I., Manzini, A., Keeling, G., Hendricks, L. A., Rieser, V., Iqbal, H., Tomašev, N., Ktena, I., Kenton, Z., Rodriguez, M., El-Sayed, S., Brown, S., Akbulut, C., Trask, A., Hughes, E., Bergman, A. S., Shelby, R., Marchal, N., Griffin, C., … Manyika, J. (2024). *The Ethics of Advanced AI Assistants* (Version 2). arXiv. https://doi.org/10.48550/ARXIV.2404.16244
14. Grancia, M. K. (2024). Decolonizing AI ethics in Africa's healthcare: An ethical perspective. *AI and Ethics*. https://doi.org/10.1007/s43681-024-00650-z
15. Grasswick, H. (2017). Epistemic Injustice in Science. In *The Routledge Handbook of Epistemic Injustice* (1st ed., pp. 313–323). Routledge.
16. Gwagwa, A., Kazim, E., & Hilliard, A. (2022). The role of the African value of Ubuntu in global AI inclusion discourse: A normative ethics perspective. *Patterns*, *3*(4), 100462. https://doi.org/10.1016/j.patter.2022.100462
17. Gwagwa, A., & Mollema, W. J. T. (2024). How could the United Nations Global Digital Compact prevent cultural imposition and hermeneutical injustice? *Patterns*, *5*(11), 101078. https://doi.org/10.1016/j.patter.2024.101078
18. Harris, K. R. & Philosophy Documentation Center. (2022). Epistemic Domination. *Thought: A Journal of Philosophy*, *11*(3), 134–141. https://doi.org/10.5840/tht202341317
19. Hookway, C. (2010). Some Varieties of Epistemic Injustice: Reflections on Fricker. *Episteme*, *7*(2), 151–163. https://doi.org/10.3366/epi.2010.0005
20. Humphreys, P. (2009). The philosophical novelty of computer simulation methods. *Synthese*, *169*(3), 615–626. https://doi.org/10.1007/s11229-008-9435-2
21. Kay, J., Kasirzadeh, A., & Mohamed, S. (2024). *Epistemic Injustice in Generative AI* (Version 1). arXiv. https://doi.org/10.48550/ARXIV.2408.11441
22. Kidd, C., & Birhane, A. (2023). How AI can distort human beliefs. *Science*, *380*(6651), 1222–1223. https://doi.org/10.1126/science.adi0248

23. Kraft, A., & Soulier, E. (2024). Knowledge-Enhanced Language Models Are Not Bias-Proof: Situated Knowledge and Epistemic Injustice in AI. *The 2024 ACM Conference on Fairness, Accountability, and Transparency*, 1433–1445. https://doi.org/10.1145/3630106.3658981
24. Lobo, C. (2022). Speaking Silences: A Wittgensteinian Inquiry into Hermeneutical Injustice. *Nordic Wittgenstein Review*. https://doi.org/10.15845/nwr.v11.3643
25. Mbembe, A. (2022). *The Earthly Community: Reflections on the Last Utopia* (Trans. Corcoran, S.). V2_ Lab for the Unstable Media.
26. McQuillan, D. (2022). *Resisting AI*. Bristol University Press.
27. McQuillan, D. (2023, June 2). *We come to bury ChatGPT, not to praise it*. https://danmcquillan.org/chatgpt.html
28. Medina, J. (2017). Varieties of Hermeneutical Injustice. In *The Routledge Handbook of Epistemic Injustice* (1st ed., pp. 41–53). Routledge.
29. Mhlambi, S., & Tiribelli, S. (2023). Decolonizing AI Ethics: Relational Autonomy as a Means to Counter AI Harms. *Topoi*, *42*(3), 867–880. https://doi.org/10.1007/s11245-022-09874-2
30. Mihalcea, R., Ignat, O., Bai, L., Borah, A., Chiruzzo, L., Zhijing, J., Kwizera, C., Nwatu, J., Poria, S., and Solorio, T. (2024). Why AI Is WEIRD and Should Not Be This Way: Towards AI For Everyone, With Everyone, By Everyone. arXiv. https://doi.org/10.48550/arXiv.2410.16315.
31. Milano, S., & Prunkl, C. (2024). Algorithmic profiling as a source of hermeneutical injustice. *Philosophical Studies*. https://doi.org/10.1007/s11098-023-02095-2
32. Mills, C. (1997). *The Racial Contract*. Cornell University Press.
33. Miragoli, M. (2024). Conformism, Ignorance & Injustice: AI as a Tool of Epistemic Oppression. *Episteme*, 1–19. https://doi.org/10.1017/epi.2024.11
34. Mohamed, S., Png, M.-T., & Isaac, W. (2020). Decolonial AI: Decolonial Theory as Sociotechnical Foresight in Artificial Intelligence. *Philosophy & Technology*, *33*(4), 659–684. https://doi.org/10.1007/s13347-020-00405-8
35. Mollema, W. J. T. (2024a). Decolonial AI as Disenclosure. *Open Journal of Social Sciences*, *12*(02), 574–603. https://doi.org/10.4236/jss.2024.122032
36. Mollema, W. J. T. (2024b). *On certainty*, Left Wittgensteinianism and conceptual change. *Theoria*, theo.12558. https://doi.org/10.1111/theo.12558
37. Moyal-Sharrock, D. (2016). Wittgenstein Today. *Wittgenstein-Studien*, *7*(1), 1–14. https://doi.org/10.1515/witt-2016-0103
38. Muldoon, J., & Wu, B. A. (2023). Artificial Intelligence in the Colonial Matrix of Power. *Philosophy & Technology*, *36*(4), 80. https://doi.org/10.1007/s13347-023-00687-8
39. Okeja, U. (2022). *Deliberative Agency: A Study in Modern African Political Philosophy*. Indiana University Press.
40. Pasquinelli, M. (n.d.). Three Thousand Years of Algorithmic Rituals: The Emergence of AI from the Computation of Space. *E-Flux Journal*, *101*. https://www.e-flux.com/journal/101/273221/three-thousand-years-of-algorithmic-rituals-the-emergence-of-ai-from-the-computation-of-space/
41. Pettit, P. (2012). *On the People's Terms: A Republican Theory and Model of Democracy* (1st ed.). Cambridge University Press. https://doi.org/10.1017/CBO9781139017428

42. Pohlhaus, G. (2017). Varieties of Epistemic Injustice. In *The Routledge Handbook of Epistemic Injustice* (1st ed., pp. 13–26). Routledge.
43. Pozzi, G. (2023). Automated opioid risk scores: A case for machine learning-induced epistemic injustice in healthcare. *Ethics and Information Technology*, *25*(1), 3. https://doi.org/10.1007/s10676-023-09676-z
44. Ricaurte, P. (2019). Data Epistemologies, The Coloniality of Power, and Resistance. *Television & New Media*, *20*(4), 350–365. https://doi.org/10.1177/1527476419831640
45. Slattery, P., Saeri, A. K., Grundy, E. A. C., Graham, J., Noetel, M., Uuk, R., Dao, J., Pour, S., Casper, S., & Thompson, N. (2024). *The AI Risk Repository: A Comprehensive Meta-Review, Database, and Taxonomy of Risks From Artificial Intelligence* (Version 1). arXiv. https://doi.org/10.48550/ARXIV.2408.12622
46. Symons, J., & Alvarado, R. (2022). Epistemic injustice and data science technologies. *Synthese*, *200*(2), 87. https://doi.org/10.1007/s11229-022-03631-z
47. Varshney, K. R. (2023). *Decolonial AI Alignment: Openness, Viśe\d{s}a-Dharma, and Including Excluded Knowledges* (Version 3). arXiv. https://doi.org/10.48550/ARXIV.2309.05030
48. Vrousalis, N. (2013). Exploitation, Vulnerability, and Social Domination. *Philosophy & Public Affairs*, *41*(2), 131–157. https://doi.org/10.1111/papa.12013
49. Vrousalis, N. (2018). Exploitation: A primer. *Philosophy Compass*, *13*(2), e12486. https://doi.org/10.1111/phc3.12486
50. Wanderer, J. (2017). Varieties of Testimonial Injustice. In *The Routledge Handbook of Epistemic Injustice* (1st ed., pp. 27–40). Routledge.
51. Wiredu, K. (2002). Conceptual decolonization as an imperative in contemporary African philosophy: Some personal reflections: *Rue Descartes*, *n° 36*(2), 53–64. https://doi.org/10.3917/rdes.036.0053
52. Wittgenstein, L. (1975). *On Certainty* (Anscombe, G. E. M. & von Wright, G. H. (eds.). Trans. Paul, D.). Blackwell.
53. Young, I. M. (1990). *Justice and the Politics of Difference*. Princeton University Press.
54. Katz, Y. (2020). *Artificial Whiteness.* Columbia University Press.
55. Bacevic, J. (2023). Epistemic injustice and epistemic positioning: towards an intersectional political economy. *Current Sociology* 71(6): 1122–1140.
56. Floridi, L. (2023). On the Brussels-Washington Consensus About the Legal Definition of Artificial Intelligence. *Philosophy & Technology* 36: 87. https://doi.org/10.1007/s13347-023-00690-z.
57. Alvarado, R. (2023). AI as an Epistemic Technology. *Science and Engineering Ethics* 29: 32. https://doi.org/10.1007/s11948-023-00451-3.
58. Báez-Vizcaíno, K. & Santana-Soriano, E. (2024). Pre-transaccional, transaccional y post- transaccional: categorías para reconocer la injusticia epistémica. *Ciencia y Sociedad, 49* (3), 9-32. https://doi.org/10.61780/cys.2024.v49i3.3235.
59. Crawford, K. (2021). *Atlas of AI: Power, Politics and the Planetary Costs of Artificial Intelligence*. Yale University Press. https://yalebooks.yale.edu/book/9780300264630/atlas-of-ai/.
60. Perrigo, B. (January 18 2023). Exclusive: OpenAI Used Kenyan Workers on Less Than $2 Per Hour to Make ChatGPT Less Toxic. Time. https://time.com/6247678/openai-chatgpt-kenya-workers/.
61. Criado Perez, C. (2019). *Invisible Women: Exposing Data Bias in a World Designed for Men.* Vintage.

62. Gurfinkel, J. (March 2023). AI and the American smile, https://medium.com/@socialcreature/ai- and-the-american-smile-76d23a0fbfaf,
63. Wittgenstein, L. (2009). *Philosophical Investigations* (Trans. Anscombe, G. E. M., Hacker, P. M. S. & Schulte, J.). Wiley-Blackwell.
64. Floridi, L. (2025). AI as Agency without Intelligence: On Artificial Intelligence as a New Form of Artificial Agency and the Multiple Realisability of Agency Thesis. *Philosophy & Technology* 38: 30. https://doi.org/10.1007/s13347-025-00858-9 .
65. Mollema, W. J. T. (2024c). Social AI and the Equation of Wittgenstein's Language User with Calvino's Literature Machine. *International Review of Literary Studies* 6(1): 39-55. https://irlsjournal.com/index.php/Irls/article/view/54.
*66.* Cappelen, H. & Denver, J. (2025). *Going Whole Hog: A Philosophical Defense of AI Cognition.* https://doi.org/10.48550/arXiv.2504.13988.
67. Dotson, K. (2014). Conceptualizing Epistemic Oppression. *Social Epistemology 28*(2): 115–138. https://doi.org/10.1080/02691728.2013.782585\
68. Wingo, A. (2024). Akan Philosophy of the Person. *The Stanford Encyclopedia of Philosophy (Summer 2024 Edition*), (eds. Zalta, E. N. & Nodelman, U.*).* https://plato.stanford.edu/archives/sum2024/entries/akan-person/.
69. The Cultural Encyclopaedia. An Essay on African Philosophical Thought. https://culturalencyclopaedia.org/an-essay-on-african-philosophical-thought-article.
70. Barnes, C. L. (2016). The Social Nature Of Individual Self-Identity: Akan And Narrative Conceptions Of Personhood. *Comparative Philosophy* 7(1): 1-19.
71. Baldwin, J. (1998). If Black English Isn't a Language, Then Tell Me, What Is? In T. Morrison (Ed.) *Collected Essays* (pp. 780-783). The Library of America.
72. Wittgenstein, L. (2009). *Philosophical Investigations.* Trans. Anscombe, G. E. M., Hacker, P. M. S. and Schulte, J. Wiley-Blackwell.